\documentclass[conference]{IEEEtran}

\usepackage{graphicx}
\usepackage{caption}
\usepackage{subcaption}
\usepackage{todonotes}

\emergencystretch=\maxdimen
\begin{document}

\IEEEoverridecommandlockouts

\title{Efficient Endangered Deer Species Monitoring with UAV Aerial Imagery and Deep Learning} 

\makeatletter 
\newcommand{\linebreakand}{%
  \end{@IEEEauthorhalign}
  \hfill\mbox{}\par
  \mbox{}\hfill\begin{@IEEEauthorhalign}
}
\makeatother 

\author{\IEEEauthorblockN{Agustín Roca}
\IEEEauthorblockA{\textit{Laboratorio de Inteligencia Artificial y Robótica}\\
\textit{Universidad de San Andrés}\\
Buenos Aires, Argentina\\
aroca@udesa.edu.ar}
\and
\IEEEauthorblockN{Gabriel Torre}
\IEEEauthorblockA{\textit{LINAR - Instituto de Ingeniería Biomédica}\\
\textit{Universidad de San Andrés - UBA}\\
Buenos Aires, Argentina\\
torreg@udesa.edu.ar}

\linebreakand

\IEEEauthorblockN{\hspace{0.5cm}Juan I. Giribet}
\IEEEauthorblockA{\textit{\hspace{0.5cm}Laboratorio de Inteligencia Artificial y Robótica}\\
\textit{\hspace{0.5cm}Universidad de San Andrés - CONICET}\\
\hspace{0.5cm}Buenos Aires, Argentina\\
\hspace{0.5cm}jgiribet@udesa.edu.ar}
\and
\IEEEauthorblockN{\hspace{0.5cm}Gastón Castro}
\IEEEauthorblockA{\hspace{0.5cm}\textit{Laboratorio de Inteligencia Artificial y Robótica}\\
\textit{\hspace{0.5cm}Universidad de San Andrés - CONICET}\\
\hspace{0.5cm}Buenos Aires, Argentina\\
\hspace{0.5cm}gcastro@udesa.edu.ar}

\linebreakand

\IEEEauthorblockN{\hspace{0.7cm}Leonardo Colombo}
\IEEEauthorblockA{\textit{\hspace{0.7cm}Centro de Automática y Robótica} \\
\textit{\hspace{0.7cm}Consejo Superior de Investigaciones Científicas}\\
\hspace{0.7cm}Madrid, España\\
\hspace{0.7cm}leonardo.colombo@csic.es}
\and
\IEEEauthorblockN{\hspace{0.7cm}Ignacio Mas}
\IEEEauthorblockA{\textit{\hspace{0.7cm}Laboratorio de Inteligencia Artificial y Robótica}\\
\textit{\hspace{0.7cm}Universidad de San Andrés - CONICET}\\
\hspace{0.7cm}Buenos Aires, Argentina\\
\hspace{0.7cm}imas@udesa.edu.ar}

\linebreakand

\IEEEauthorblockN{Javier Pereira}
\IEEEauthorblockA{\textit{Museo Argentino de Ciencias Naturales "Bernardino Rivadavia"} \\
\textit{CONICET}\\
CABA, Argentina \\
jpereira@conicet.gov.ar}
}

\maketitle

\vspace{-1em}
\noindent
\begin{center}
    \parbox{0.95\linewidth}{
        \footnotesize
        © 2024 IEEE. Personal use of this material is permitted.  Permission from IEEE must be obtained for all other uses, in any current or future media, including reprinting/republishing this material for advertising or promotional purposes, creating new collective works, for resale or redistribution to servers or lists, or reuse of any copyrighted component of this work in other works.
    }
\end{center}
\vspace{0.5em}

\thispagestyle{empty}
\pagestyle{empty}

\begin{abstract}
This paper examines the use of Unmanned Aerial Vehicles (UAVs) and deep learning for detecting endangered deer species in their natural habitats. As traditional identification processes require trained manual labor that can be costly in resources and time, there is a need for more efficient solutions. Leveraging high-resolution aerial imagery, advanced computer vision techniques are applied to automate the identification process of deer across two distinct projects in Buenos Aires, Argentina. The first project, Pantano Project, involves the marsh deer in the Paraná Delta, while the second, WiMoBo, focuses on the Pampas deer in Campos del Tuyú National Park. A tailored algorithm was developed using the YOLO framework, trained on extensive datasets compiled from UAV-captured images. The findings demonstrate that the algorithm effectively identifies marsh deer with a high degree of accuracy and provides initial insights into its applicability to Pampas deer, albeit with noted limitations. This study not only supports ongoing conservation efforts but also highlights the potential of integrating AI with UAV technology to enhance wildlife monitoring and management practices.

\end{abstract}


\section{Introduction}
Monitoring wildlife plays an essential role in understanding and preserving ecosystems. Traditional methods of wildlife observation, however, can be labor-intensive, costly, and limited in scope \cite{buckland2004advanced, groom2013remote}. The advent of Unmanned Aerial Vehicles (UAVs) offers a promising alternative, providing expansive area coverage and the ability to gather data remotely. Despite these advantages, challenges remain in the effective detection and identification of small or camouflaged wildlife species such as deer, through aerial imagery. This work analyzes the application of artificial intelligence techniques for the automatic identification of deer in aerial images, thus addressing a critical challenge in wildlife conservation.

This section introduces the Pantano Project and WiMoBo Project, detailing their objectives and significance. Section \ref{section:methodology} discusses the methodology, including data acquisition, processing, and the application of the YOLO framework for deer detection. Section \ref{section:results} presents the results, evaluating the performance of the detection algorithm and its application to different deer species. Section \ref{section:future_directions} outlines future directions for improving the detection system and extending its application. Finally, Section \ref{section:conclusions} concludes the paper, summarizing the key findings and their implications for wildlife conservation.

\subsection{Pantano Project}

In the environment of the Paraná Delta, home to various important species such as the marsh deer (Blastocerus dichotomus), a multidisciplinary team  belonging to the Pantano Project has been carrying out fundamental work for the protection and conservation of this endangered species. Initiated in 2014 and involving scientists from CONICET, INTA, two NGOs (ACEN and CeIBA), and independent technicians and professionals from other disciplines, the project emerged in response to the growing threats faced by marsh deer in the region. One of the challenges Pantano Project faces is understanding the ecology and behavior of marsh deer, as well as developing effective conservation strategies. The use of UAVs represents a significant milestone in this initiative, allowing for the remote detection of these cervids in their natural environment.

Recently, a census of marsh deer was conducted through the analysis of aerial images collected in the Paraná Delta. For this, work was carried out in conjunction with researchers and technicians from the Centro de Investigación Científica y de Transferencia Tecnológica a la Producción (CICYTTP – CONICET, Diamante, Entre Ríos), and 168 volunteers were trained in image analysis and species identification.

This paper presents the initial results obtained to achieve automatic detection of deer using artificial intelligence algorithms, results that will not only be useful for the preservation of marsh deer but also establish links with other ongoing projects in the region.

\subsection{WiMoBo Project}

The Pampas deer (Ozotoceros bezoarticus) is another native South American species that is endangered in Argentina. The last population of this species in Buenos Aires province finds refuge in Samborombón Bay, an area of international wetlands importance (Ramsar site) since 1997. The population decline and change in the distribution of the Pampas deer are mainly attributed to negative interactions with various species introduced by humans such as wild pigs and packs of dogs.

Campos del Tuyú National Park (PNCDT), covering 3,040 hectares within the bay, has as one of its main objectives the conservation of a population of deer and Pampas grassland. However, the floodable nature of PNCDT prevents access with ground vehicles, making regular censuses difficult to control the population and protect it against invasive predators. The proximity of the park to the fishing port of General Lavalle adds urgency to environmental controls in the area.

In this context, the Wildlife Monitoring Bots (WiMoBo) project emerges, a Spanish-Argentine research network consolidated through the LINCGLOBAL 2022 project, integrated by scientists from the CSIC, CONICET, Argentine National Parks Administration, the Vida Silvestre Foundation, Universidad Nacional de La Plata, Universidad de Buenos Aires, and the Universidad de San Andrés. The main goal of WiMoBo is to develop a multi-domain and multi-robot system for the population monitoring of Pampas deer in the PNCDT. This project seeks to strengthen the development of navigation and control algorithms, allowing the collaboration between autonomous vehicles in different domains (air and water) to perform environmental monitoring tasks and preserve the endangered species along with its ecosystem.

The research group brings together mathematicians, biologists, and engineers from Argentina and Spain, creating an interdisciplinary environment conducive to addressing the challenges in the conservation of threatened species and their ecosystems.

\subsection{Objectives}
In the context of the aforementioned projects, various scientific challenges arise, including the automatic detection of the species to be monitored, through the use of aerial images taken from UAVs. This work presents the first results in this direction.

Using tools like YOLO \cite{yolov8_ultralytics} and the image database generated within the framework of Pantano Project, an algorithm was trained that allows the automatic detection of deer. The results are contrasted with a census carried out by experts and trained volunteers who reviewed each of the photos to detect the presence of marsh deer. The performance of the algorithm for automatic detection is analyzed, with the goal of, in the future, automating the task of detection and census of the species.

On the other hand, the performance of the trained network using the information obtained in Pantano Project, for the detection of Pampas deer, is analyzed. Although they are dissimilar species in various respects, as well as the environment in which they are found, it is interesting to analyze how the network behaves in the detection of Pampas deer, since this will allow decisions to be made about the fieldwork to be carried out in the PNCDT.

\section{Methodology}
\label{section:methodology}

A structured approach was adopted, encompassing the acquisition and processing of data, as well as the implementation of advanced deep learning techniques for wildlife detection.

\subsection{Related Work}
Due to various factors, such as the increase in poaching of emblematic species and the alteration of natural environments, more and more species are in danger. The rapid loss of habitat and environmental degradation exacerbate these circumstances, making regular monitoring of wildlife essential to understand the processes in which the pace surpasses the usual changes in wildlife communities. With the aim of simplifying the task of observation and reducing costs, UAVs have been used for this task for several years.

UAVs have been used in wildlife observation for more than a decade. In \cite{Linchant2015}, for example, the main advances and challenges in the use of UAVs for these applications are detailed. This work compiles the main results up to the year 2015, and among its conclusions, it highlights that UAVs have technical limitations, such as the inability to cover large geographical areas, and therefore, it is necessary to create and evaluate new protocols for sampling methods, inventories, and statistical analysis. The same work indicates that the use of thermal sensors combined with high-definition optical images is expected to increase species discrimination. Additionally, the study concludes that one of the main limitations is the lack of regulations that allow the use of this technology with UAVs.

Since 2015 to date, great advances have been made, both in regulation (albeit more slowly), as well as in the design of methodologies for sampling and statistical analysis \cite{fust2023increasing, pereira2023}, as well as in the use of sensor fusion techniques for the discrimination of various species. In particular, in recent years, automatic detection strategies using artificial intelligence algorithms have proven to be a useful tool to facilitate the task of detection and classification of species.

Various methods have been investigated for detecting wildlife in aerial images, which include target detection algorithms, semantic segmentation algorithms, and deep learning methods
\cite{chabot2016computer, li2019use, sundaram2020fsscaps}.
For example, Barbedo et al. used convolutional neural networks to monitor cattle herds with images acquired with UAVs \cite{barbedo2019study}. In \cite{brown2022automated}, high-resolution UAV images were used to establish an accurate field truth, using a target detection-based method with YOLOv5 to achieve localization and counting of animals on farms in Australia. Deep learning models, such as YOLO, have emerged as a significant approach in UAV-based aerial image applications for wildlife detection.

The success of deep learning in UAV wildlife detection depends on having a large amount of real-world data available to train algorithms \cite{weiss2016survey}. Due to the lack of large-scale wildlife aerial image datasets, most current methods involve adapting object detection algorithms developed for natural scene images to aerial images, which is not suitable for wildlife detection \cite{zheng2021self}. These challenges contribute to the current shortcomings of UAV-based wildlife detection algorithms, including low accuracy, poor robustness, and unsatisfactory results in practical application \cite{okafor2017operational,kellenberger2018best}.

In \cite{Mou2023}, a real-time species detection model based on YOLOv7 is presented, adapted to improve the detection of small targets in UAV wildlife monitoring. Additionally, the Wildlife Aerial Images from Drone (WAID) dataset is introduced, which includes 14,375 high-quality UAV aerial images under different environmental conditions, covering six wildlife species and multiple habitat types. The study includes statistical analysis experiments, algorithm comparison, and dataset generalization, demonstrating that WAID is suitable for UAV wildlife monitoring algorithm research, and that SE-YOLO is the most effective method, with a mAP of up to 0.983, and in particular a mAP of 0.926 for animals that occupy less than 50 pixels in the image. In \cite{Lyu2024}, information from an RGB camera and a thermal camera is used for deer detection. In particular, the authors show that it is difficult to adapt a network trained with visible images to work correctly in the thermal spectrum, particularly because these images usually do not incorporate as detailed information as RGB images, but with a well-trained network architecture, the authors show that performance in deer detection significantly increases.

In this work, an approach using the YOLOv8 model is employed for the detection of marsh deer in UAV-acquired images. This method specifically focuses on detecting marsh deer, but the used dataset also includes images with other animals such as cows, capybaras, and birds, demonstrating the model's ability to handle diverse wildlife in the detection process. This approach represents a significant advancement in UAV-based wildlife monitoring, combining deep learning techniques with real-world applications to enhance conservation efforts.

\subsection{Data Acquisition}

A crucial point for the use of certain tools for the automatic detection of wildlife is having a sufficiently rich database to train the algorithms. For this work, the information obtained within the framework of Pantano Project is available.

A survey was also conducted in PNCDT, and aerial images of the deer were obtained from different vehicles. Although the available database is not as extensive as that of Pantano Project, it proves useful for validating how the network trained on one species of cervid in a specific habitat generalizes its learning to another species with different characteristics and habitat.

\subsubsection{Vehicles and cameras used}
\subsubsection*{Pantano Project}
For collecting the data used to create the database in the framework of Pantano Project, Phantom 4 Pro UAV quadrotors were used, each equipped with a high-definition 20 MP camera with an 84° field of view, 8.8/24 mm lens, auto-focus from f/2.8 to f/11 at 1m mounted on its three-axis stabilizer. The vehicle performed flights automatically, loading the flight route into DJI's proprietary software, and once the UAV was launched, it followed the pre-programmed route from the starting point of the flight, with a ground operator observing remotely.

\subsubsection*{WiMoBo Project}
In the case of the information collected in the Campos del Tuyú National Park, it was acquired using fixed-wing vehicles and a hexa-rotor vehicle built in the laboratory, and a DJI Mavic Pro 2 quadrotor was also used. The fixed-wing vehicles have a wingspan of 980mm and a length of 450mm,
and are equipped with a 2216-1400 KV motor. The propeller used is 8x5. The power comes from a 4-cell battery with a capacity of 3500 mAh. To control the motor speed, it has a 40A ESC.
The minimum and maximum flight speeds are 31km/h and 130 km/h, respectively, while the cruising speed is around 80km/h. The vehicle's weight is 860g-900g, depending on the selected battery.
The fixed-wing vehicle is equipped with a GoPro Hero5 Session camera, which captures video in 4K, 2.7K, 1440p, and 1080p and takes 10MP photos at 30 fps. Although the vehicle was flown in automatic mode, the pilot has the possibility at any time to take control of the vehicle, for which it has a control and telemetry link at 433MHz, with a range of 50km. On the other hand, real-time video transmission is available via a 5.8G link.

\begin{figure}[h]
\centering
\includegraphics[width=\linewidth]{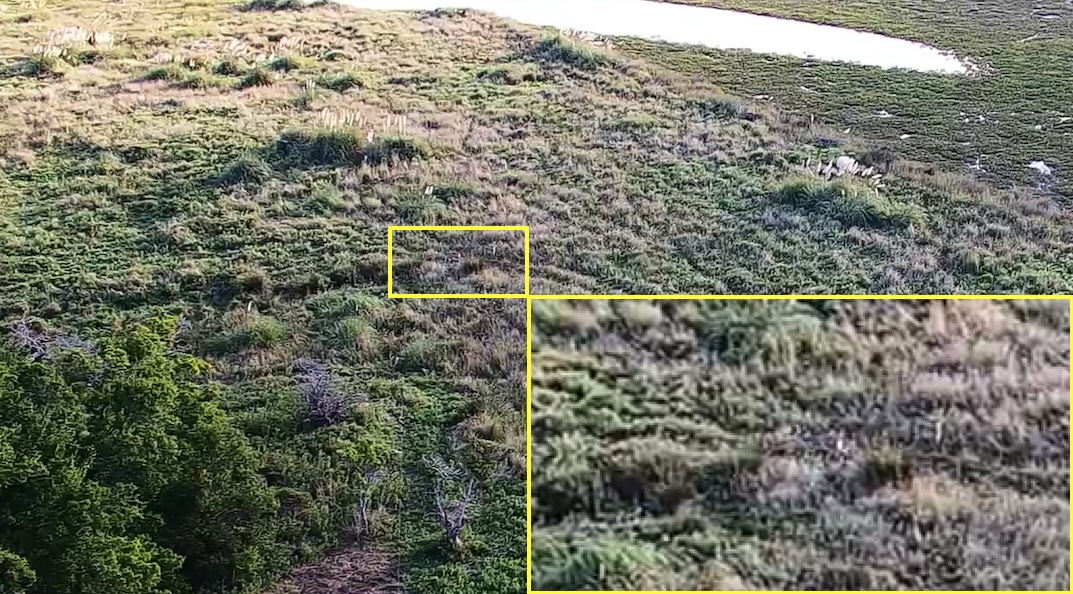}
\caption{Video images taken from the hexarotor at 50m. The yellow area shows an enlarged zone where two deer are visible.}
\label{fig:venado-rgb}
\end{figure}

The hexarotor has 6 T-Motor MN4014 400KV motors with a 40A ESC, which with 6s 10000 mAh batteries and 15x5 propellers achieve a thrust of 2.5kg. The hexarotor's frame is made of carbon fiber, with a maximum distance between motors of 850mm and a weight of approximately 1210g. The arms are retractable, to facilitate transport. This vehicle also used a radio control and telemetry at 433MHz.

The hexarotor is equipped with an RGB and thermal camera, model TH10T6LN, with a 10x optical zoom and a thermal resolution of 640x512, along with a compatible laser distance measurement system for distances up to 1500 meters. The camera has a three-axis stabilizing gimbal. In addition, it supports IP HD 1080P output at 30fps with a 4 megapixel CMOS sensor. In this work, the information obtained by the thermal camera was not included for training the network, because there are not many thermal images of the deer. However, campaigns are expected to be conducted to fly in more areas and obtain more thermal images of the deer, as the results seem promising.

In Figures \ref{fig:venado-rgb} and \ref{fig:venado-termica}, a comparison is presented between images captured in one of the flights at PNCDT, by the hexarotor's camera in the visible spectrum and in the thermal spectrum. In Fig. \ref{fig:venado-rgb}, the detection of the deer is practically impossible, even when zooming into the image, obtained as a frame from a 1080p video. In contrast, the same scene is captured by the thermal video, as shown in Fig. \ref{fig:venado-termica}, where the silhouette of two deer is clearly visible.

It is possible to detect the deer in the RGB video, due to their movement, but not when analyzing a frame. In certain cases, dynamic information can be useful for detecting animals, but ideally, the deer should not be disturbed by the presence of the UAV, and therefore most of the videos obtained may not have enough information to easily detect the deer.

\begin{figure}[h]
\centering
\includegraphics[width=\linewidth]{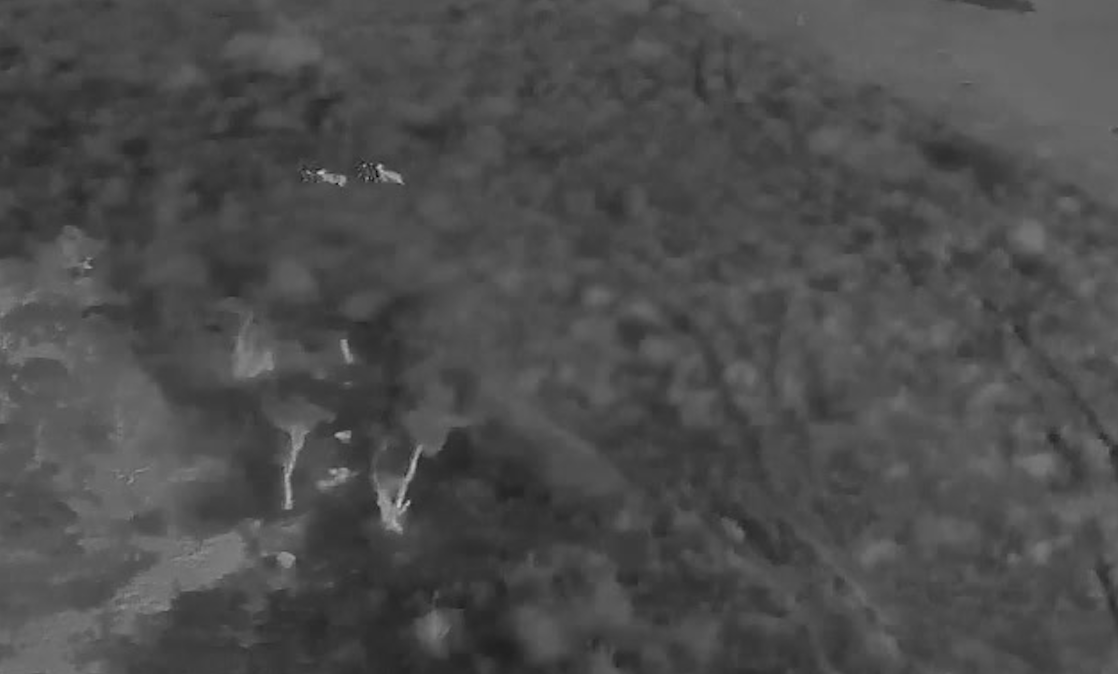}
\caption{Thermal video images from the hexarotor at 50m. The silhouette of two deer can be seen.}
\label{fig:venado-termica}
\end{figure}

The autopilot of both vehicles is a Pixhawk v2.4.8, with PX4 firmware configured for each of the vehicles. The QGroundControl software is used to configure the vehicles and program the missions.

\subsubsection{Routes}
\subsubsection*{Pantano Project}
An initial exploratory study was conducted on May 16 and 17, 2019, during which six flight plans were executed (totaling 19.5 km). These flights were designed to test different flight parameters (i.e., speed, altitude) and image collection times to maximize the probability of distinguishing a marsh deer from its environment in the various habitat types within the study area. Based on the analysis performed, an altitude of 45 m above ground level (resulting in a transect 67.5 m wide), a speed of 6.5 m/s, and a photograph taken every 5 seconds, with approximately a 33\% front overlap between consecutive photographs, were established.
No escape behavior from the deer was observed at the advance of a UAV for the chosen trajectory parameters.

The study flights were conducted on August 6 and 8, 2019, in winter, given that the leaf cover is lower in the poplars and willows present in the deer's habitat. A grid with 1500m (north-south) x 100m (east-west) cells was superimposed on an image of the study area, and 1500m transects were defined by the intersection of north-south lines with east-west lines. As the perimeter of the study area is irregular and the length of the resulting transects on the edges varied considerably, only transects longer than 760m were used. The transects were numbered and randomly selected to be included in the study until a coverage of 10\% of the study area was reached. From the selected transects, flight routes were set using DJI GS Pro software. Each flight route was designed to cover as many chosen transects as possible, considering the flight range limitations imposed by battery capacity. Once these flight routes were defined, additional transects were included. When the start and end positions of the transects were separated by more than 1500m, additional east-west transects were delimited to be traversed. The first and last 100m of these transects were truncated. That is, the photographs were discarded to avoid a possible double count of individuals when the count by experts was made. However, these images were included for training the algorithm, since they contained useful information.

Flight routes were executed consecutively separated by 500m within a short period of time (less than 3 hours), to minimize double counting, since the deer could have moved between adjacent transects.

\subsubsection*{WiMoBo Project}
The first flights at PNCDT took place in late December 2022. During this initial phase, a series of flights were carried out using quadrotors, particularly with the DJI vehicle, trials were conducted at various altitudes to observe the deer's behavior in the presence of the UAV. The main objective was to determine an optimal flight altitude that would not disturb the deer. From observations of their behavior, it was concluded that a minimum altitude of 25 meters above sea level was necessary to prevent the deer from feeling threatened by the presence of the UAV. However, it is important to note that this initial analysis is specific to the DJI quadcopter and may not be easily applicable to other types of aerial vehicles.

In February 2024, a second visit to PNCDT was made with the main goal of capturing thermal images, as well as videos taken from a fixed-wing vehicle. During this visit, additional images were also obtained from the DJI quadcopter, which proved useful for evaluating the effectiveness of a neural network trained with marsh deer data in detecting Pampas deer.

Despite these advances, there is still a need to define a methodology similar to that used in Pantano Project, as well as to identify suitable transects for conducting a survey of Pampas deer in PNCDT. It is expected that in the near future a campaign to capture images in the park can be carried out.

\subsection{Data Processing}

A total of 39,798 photographs were obtained, divided into 575 transects, in the flights carried out within the framework of Pantano Project. These were manually reviewed by 4 expert scientists and 168 trained volunteers, cataloging those in which marsh deer were seen. Each observer analyzed a subset of images, following a standardized protocol: zooming in digitally on the image and looking from left to right and top to bottom for marsh deer. Sequential overlapping images were compared to avoid double counting. Each photograph was verified by at least two independent observers. The total number of marsh deer recorded by all observers, excluding multiple detections of the same individual by more than one observer, was used to estimate the density of marsh deer.

A total of 231 deer were identified by the observers in the analyzed photographs. Additionally, during this work, 88 images containing cows, which can be mistaken for deer due to similar sizes and colors, were also identified. Three images captured other animals (two capybaras and one bird). Some of this images can be seen in Fig. \ref{fig:animales}. Importantly, the analysis also revealed an image with a deer that had not been initially labeled by the observers. Detailed results of this discovery will be discussed in the results section

\begin{figure*}[h]
\centering
\includegraphics[width=\linewidth]{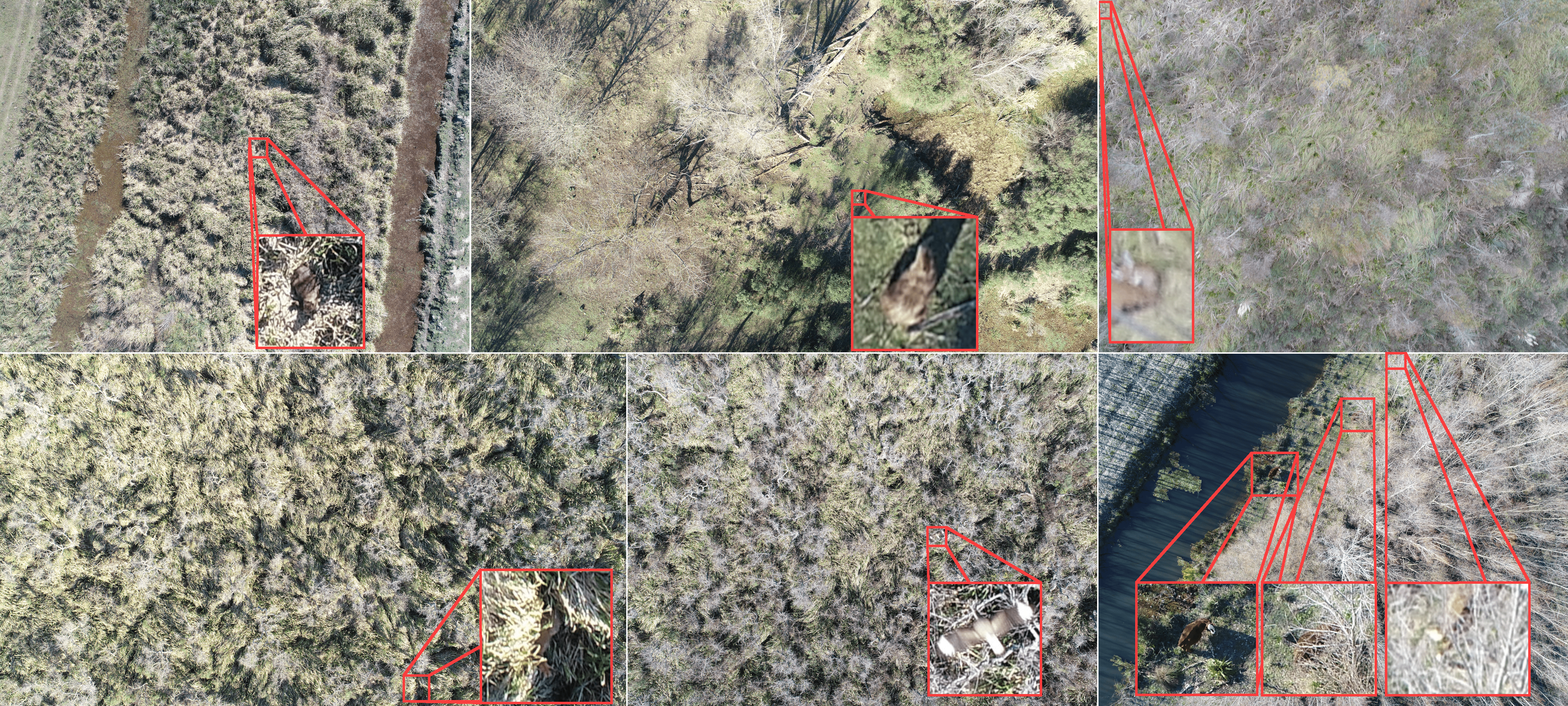}
\caption{Images with various animals found, including capybaras, deer, cows, and birds}
\label{fig:animales}
\end{figure*}

Masks were generated for the 232 positive deer images to identify specifically which pixels the deer occupies.

A fine-tuning of the YOLOv8n-seg model was performed using 140 images with deer with their respective masks, 54 images with cows, the 3 images with other animals, and 575 images without any animals (one per transect). During the training, a validation set containing 46 images with deer and 17 images with cows, and 575 images without animals (one per transect) was used.

The network was trained reducing the images to a size of 1280$\times$1280 pixels, allowing the images to be inverted and rotated at any angle. It was also specified that two instances of deer do not overlap.

The training of the model was conducted over 440 epochs, finding an optimum in the validation set at epoch 340. The entire training process took approximately 6 hours and 28 minutes (23,252 seconds) to complete. 

\section{Results}
\label{section:results}

The subsequent sections provide a comprehensive analysis of the findings, illustrating the performance and implications of the developed detection algorithm.

\subsection{Detection Evaluation}
The performance of the system was evaluated using standard object detection metrics. For all the results below, a test set containing 46 images with deer, 17 images with cows, and 575 images without animals was used. None of the images in this set were present during the training or validation of the model.

Running on a server with an Nvidia T4 GPU, the model took 3 hours and 43 minutes to perform detections for the 39,798 images. This equates to almost 3 images per second.

As a preliminary evaluation, the $mAP_{10}$ was measured at different confidence thresholds. An IoU of 10\% was chosen because the deer occupy a very small section of the image, and the perfect framing of the deer is not relevant for this problem.

\begin{figure}[h]
\centering
\includegraphics[width=\linewidth]{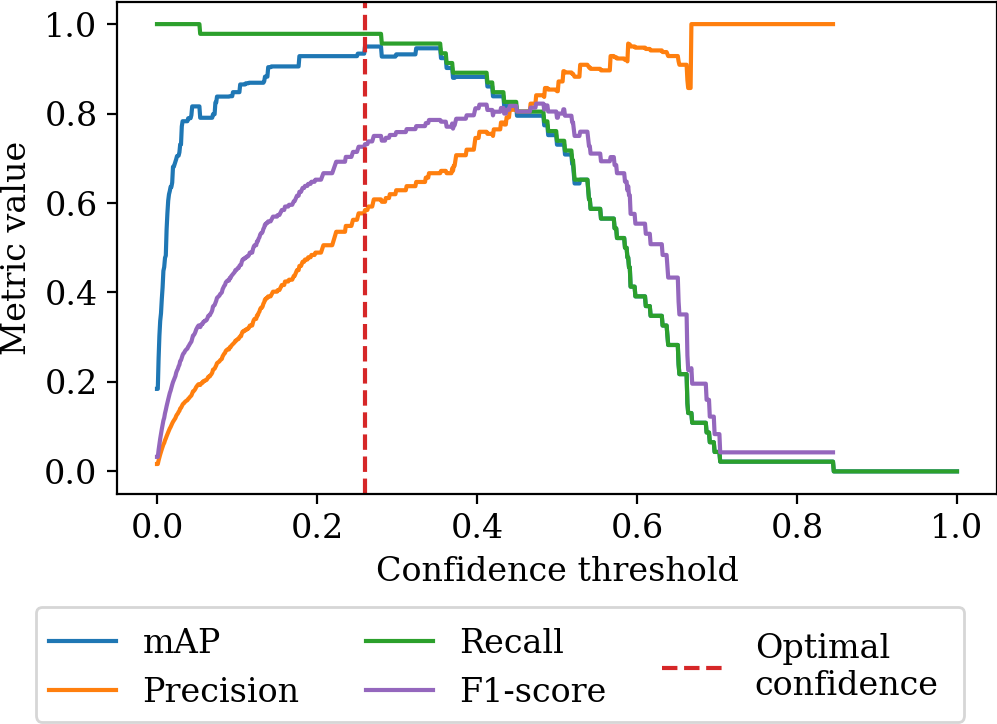}
\caption{Confusion Matrix with the optimal confidence threshold (0.260)}
\label{fig:metrics}
\end{figure}

Thus, a confidence threshold that maximizes the $mAP_{10}$ was found at 0.260, as seen in Fig. \ref{fig:metrics}. Fig. \ref{fig:mAP} shows the Precision-Recall curve with this confidence threshold.

\begin{figure}[h]
\centering
\includegraphics[width=0.9\linewidth]{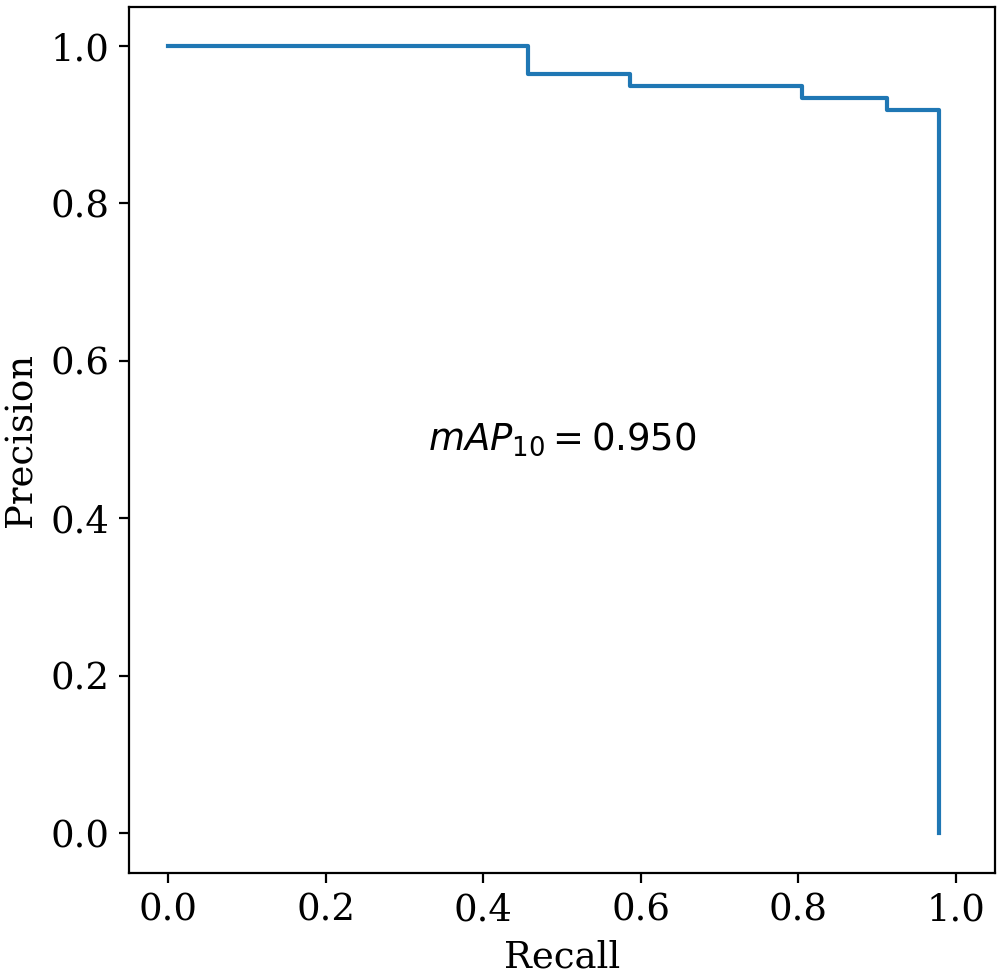}
\caption{Precision-Recall curve for an IoU threshold of 10\% with the optimal confidence threshold (0.260)}
\label{fig:mAP}
\end{figure}

\begin{figure}[h]
\centering
\includegraphics[width=\linewidth]{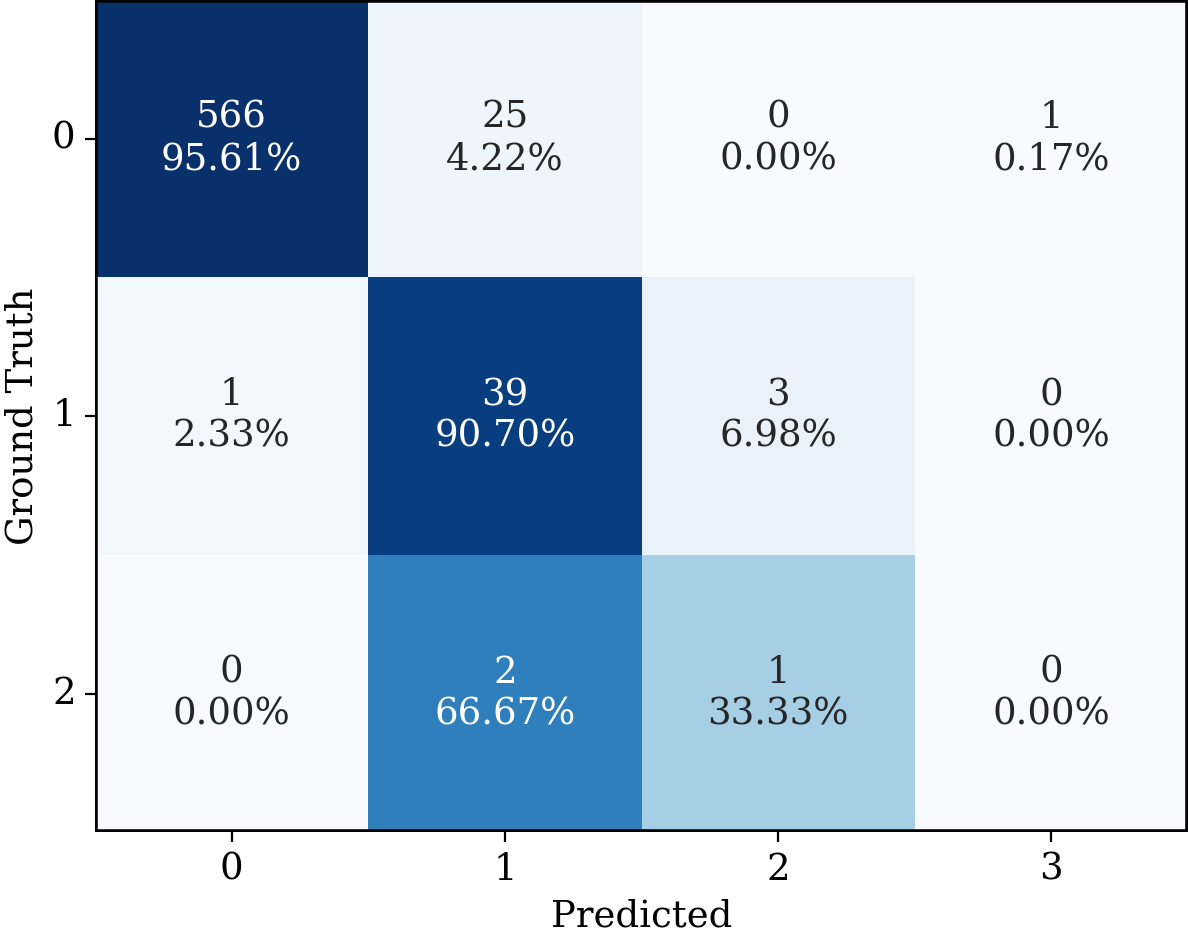}
\caption{Confusion matrix with the optimal confidence threshold (0.260)}
\label{fig:confusion_matrix}
\end{figure}

Using the selected optimal threshold, the number of deer that the model detects in each image of the test set is counted. The resulting confusion matrix can be seen in Fig. \ref{fig:confusion_matrix}.

\subsection{Finding of an Unlabeled Deer}

During the extensive review process carried out by the observers, a vast number of deer were successfully identified and cataloged. Nevertheless, the advanced capabilities of the YOLOv8-powered detection model allowed for the identification of an additional deer that had not been previously labeled, as seen in Fig. \ref{fig:lost_deer}.

This instance was discovered during the early iterations of the analysis. It underscores the utility of integrating automated detection systems to augment human observations, particularly in extensive datasets where small details may be exceptionally challenging to discern consistently.

\begin{figure}[h]
\centering
\includegraphics[width=\linewidth]{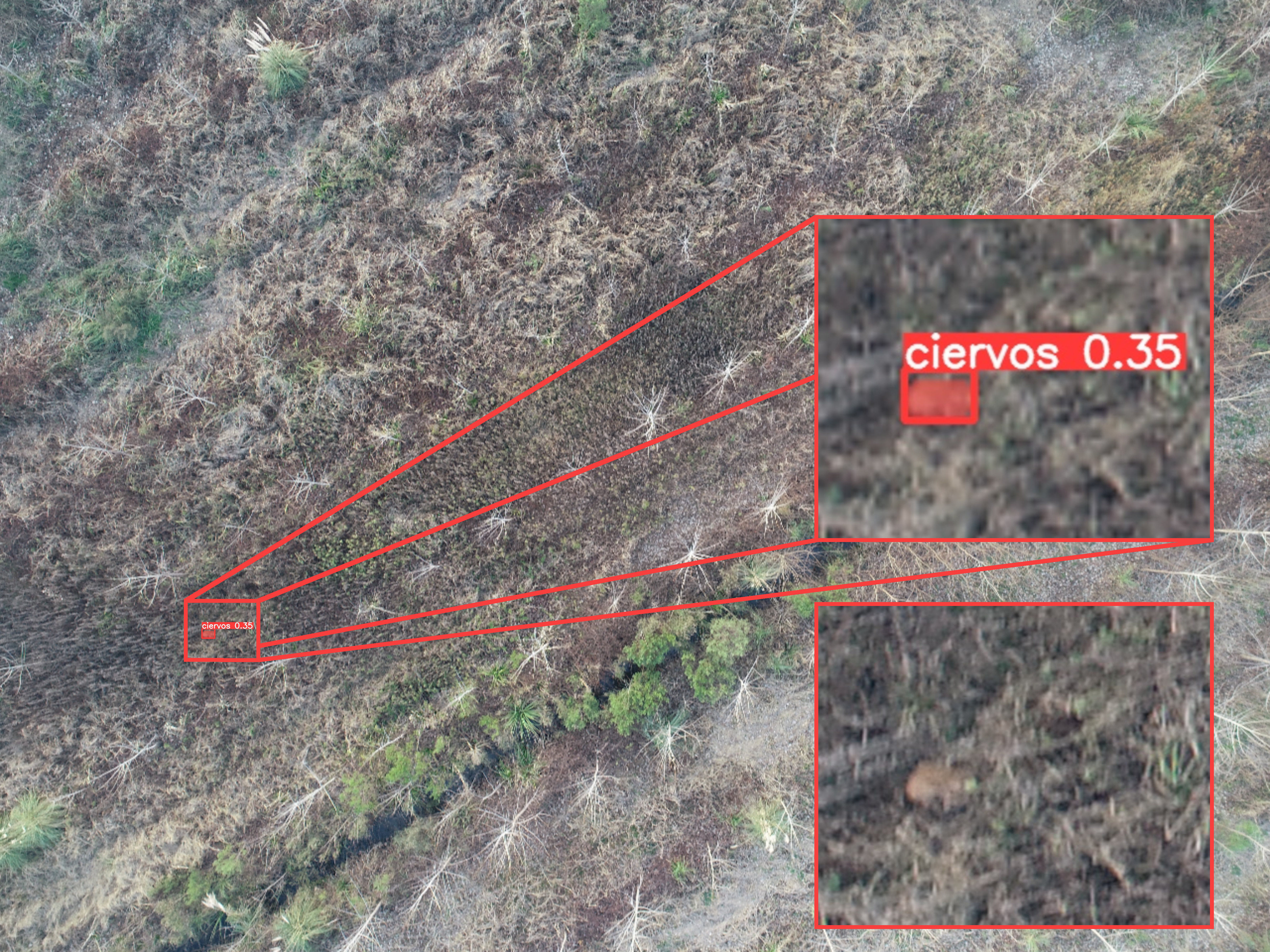}
\caption{Image with the deer that had not originally been detected by the human labelers.}
\label{fig:lost_deer}
\end{figure}

\subsection{Evaluation on Pampas Deer}

The model trained on the marsh deer database was evaluated on data from Pampas deer, which differ in coloration, particularly in the lower parts where the deer are paler, and the marsh deer are more brownish as seen in Fig. \ref{fig:venado}. Additionally, the images of the Pampas deer are not taken from a bird's eye view. The model was able to detect some Pampas deer with thresholds close to 0.1 (see Fig. \ref{fig:venado_det}), indicating that while the model is capable of identifying the Pampas deer, there are certain limitations.

\begin{figure}[h]
\centering
\includegraphics[width=\linewidth]{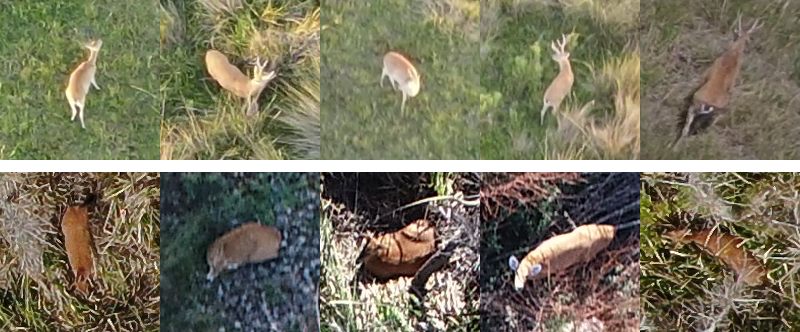}
\caption{Top: Pampas deer. Bottom: Marsh deer}
\label{fig:venado}
\end{figure}

\begin{figure}[h]
\centering
\includegraphics[width=\linewidth]{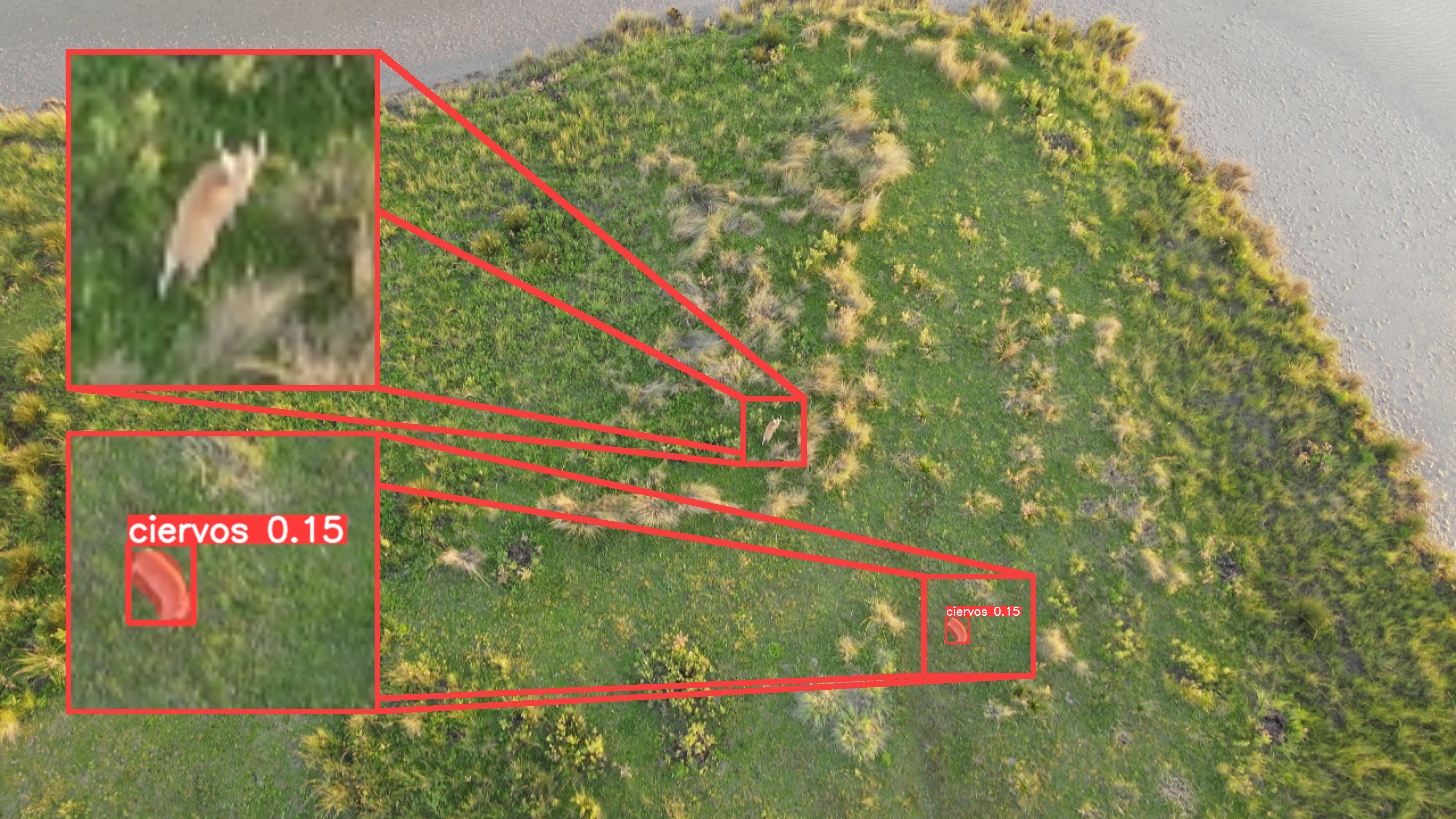}
\caption{Section of an image with a Pampas deer detected by the model and other Pampas deer that the model could not detect}
\label{fig:venado_det}
\end{figure}

\section{Future Directions}
\label{section:future_directions}
As a next step, it is planned to collect an image dataset on Pampas deer in PNCDT. Currently, an analysis of suitable sites for conducting flights is being carried out, as well as the selection of suitable vehicles and cameras for data collection.

Additionally, the integration of thermal sensors is planned in the data acquisition processes for both Pampas and marsh deer. The use of thermal imagery is expected to enhance the ability to detect cervids in various environmental conditions, especially during nighttime or in dense foliage, where visual spectrum cameras may fall short. By analyzing how thermal data complements traditional RGB imaging, the robustness and accuracy of the detection systems are anticipated to improve significantly.

\section{Conclusions}
\label{section:conclusions}
This work presents an approach for the automatic detection of endangered deer species using UAV aerial images combined with the YOLO deep learning framework. High-definition RGB images of deer, which were manually tagged by 168 trained human labelers, provided a substantial base for training and refining the detection models. The study highlights the effective application of these models in real-world conservation efforts. The results obtained show that the network reaches an $mAP_{10}$ of 0.950 and demonstrate the potential of automated deer detection in natural environments using UAVs.

The performance of the network trained with information from the marsh deer was evaluated for the detection of other cervids, specifically the Pampas deer. In this case, the network demonstrated rather poor performance, unable to detect deer unless the confidence threshold was considerably lowered.

While this may be considered a negative point, since it appears necessary to extend the database with information on Pampas deer, it also has a positive aspect because the system does not confuse Pampas deer with marsh deer. This could serve for applications in environments with multiple cervid species.

\section*{
Acknowledgments
}
{
We thank the National Parks for all the support received to carry out this research. This work has been partially funded by the project PICT-2019-0373 and PICT-2019-2371 of the ANPCyT, Argentina, and by the project LINCGLOBAL 2022, of the Spanish National Research Council.
}
\bibliographystyle{IEEEtran}
\bibliography{referencias}

\end{document}